\documentclass{article}

\usepackage{microtype}
\usepackage{graphicx}
\usepackage{subcaption}
\usepackage{booktabs}
\usepackage{bm}
\usepackage{hyperref}
\usepackage[commands, environments]{AVT}

\usepackage{algorithm}
\usepackage{algorithmic}
\usepackage[accepted]{icml2026}
\usepackage{amsmath}
\usepackage{amssymb}
\usepackage{mathtools}
\usepackage{amsthm}
\usepackage[capitalize,noabbrev]{cleveref}

\theoremstyle{plain}
\newtheorem{theorem}{Theorem}[section]

\theoremstyle{definition}

\theoremstyle{remark}
\newtheorem{remark}[theorem]{Remark}

\DeclareMathOperator{\Id}{\mathrm{Id}}
\DeclareMathOperator{\Tr}{\mathrm{Tr}}

\DeclareMathOperator{\diag}{diag}
\DeclareMathOperator{\Div}{div}
\def\abs#1{\left|#1\right|}
\def\norm#1{\left\lVert#1\right\rVert}

\def\v#1{\boldsymbol{#1}}

\hypersetup{
    colorlinks=true,
    citecolor=teal,
    linkcolor=teal,
    urlcolor=teal,
    filecolor=teal                         
}

\newcommand{\cN}{\mathcal{N}}

\DeclarePairedDelimiterX{\infdiv}[2]{(}{)}{#1\;\delimsize\|\;#2}
\usepackage[textsize=tiny]{todonotes}

\icmltitlerunning{Variance-Tilted Diffusion Models for Diverse Sampling}

\begin{document}

\twocolumn[
  \icmltitle{Variance-Tilted Diffusion Models for Diverse Sampling}

  \icmlsetsymbol{equal}{*}

  \begin{icmlauthorlist}
    \icmlauthor{Iskander Azangulov}{yyy}
    \icmlauthor{Leo Zhang}{yyy}
    \icmlauthor{Kianoosh Ashouritaklimi }{yyy}
  \end{icmlauthorlist}

  \icmlaffiliation{yyy}{Department of Statistics, University of Oxford, Oxford, UK}

  \icmlcorrespondingauthor{Iskander Azangulov}{iskander.azangulov@stats.ox.ac.uk}

  \icmlkeywords{Machine Learning, Diffusion Models, Diverse Sampling}

  \vskip 0.3in
]

\printAffiliationsAndNotice{}

\begin{abstract}
Diffusion models are typically sampled independently, even when the downstream objective is to obtain a diverse set of candidates. We introduce a variance-weighted batch distribution that favors collections of samples with large empirical spread after a prescribed linear feature map. The target is specified explicitly, and the sampler is derived as the corresponding Doob $h$-transform of independent diffusion dynamics. The resulting correction has a compact form: an interaction term that repels posterior denoised means, together with a curvature term that moves particles to the region of higher feature variance. This yields an interacting-particle sampler with a transparent probabilistic target rather than a heuristic repulsive drift. 
\end{abstract}

\section{Introduction}
Diffusion models \cite{sohl2015deep, ho2020denoising, song2020score} have established themselves as the gold standard for generative modeling, achieving unprecedented fidelity in domains ranging from high-resolution image synthesis \citep{dhariwal2021diffusion, rombach2022high} to scientific discovery, such as protein folding \citep{watson2023molecule, abramson2024accurate} and molecular docking \cite{prat2025sigmadock}. However, the standard sampling paradigm---drawing independent and identically distributed (i.i.d.) samples---is often suboptimal for downstream tasks. In many settings, the goal is not merely to generate high-probability modes, but to uncover a set of candidates that covers distinct regions of the data manifold. For instance, in de novo drug design, generating structurally diverse ligands is crucial to mitigate the risk of late-stage failure due to shared toxicity profiles. Similarly, in robotic motion planning, a diverse set of trajectory proposals provides robustness against unforeseen environmental obstacles.

To enforce such diversity, approaches such as particle guidance \citep{corso2023particleguidancenoniiddiverse} introduce repulsive potentials into the sampling dynamics. While effective at increasing diversity, these methods modify the drift heuristically, resulting in an implicit target distribution that is computationally intractable to evaluate or control precisely.

\paragraph{Contributions.}
In contrast to prior work, we define the batch target distribution first and subsequently derive the exact sampler that realizes it. For a batch of $n$ samples $\bm{x}=(x^{(1)},\ldots,x^{(n)})$, we study the variance-weighted distribution
\begin{equation}
    \pi_0^A(\bm{x})
    \propto
    \Var_n^A(\bm{x})\prod_{i=1}^n p_0(x^{(i)}),
    \label{eq:target_intro}
\end{equation}
where $A:\R^d\to\R^k$ is a linear operator that maps samples to a task-specific feature space, and the empirical variance in this space is defined as
\begin{equation}
    \Var_n^A(\bm{x})
    :=
    \frac1n\sum_{i=1}^n \norm{A(x^{(i)}-\bar x)}^2,
    \text{ with } 
    \bar x:=\frac1n\sum_{i=1}^n x^{(i)}.
    \label{eq:variance_A_intro}
\end{equation}
The linear operator $A$ dictates the representation used to encourage diversity; for example, it can select specific coordinates, isolate low-frequency Fourier coefficients, mask spatial regions, or extract other relevant linear features.

Our primary contributions are as follows:
\begin{itemize}
    \item \textbf{Target-first formulation.} We frame diverse generation as an exact sampling problem from the explicit batch target \eqref{eq:target_intro}, rather than relying on an uncontrolled repulsive force.
    \item \textbf{Closed-form Doob correction.} We compute the Doob $h$-transform induced by the variance tilt and provide the corresponding score correction for a general linear operator $A$.
    \item \textbf{Tractable sampling algorithm.} We introduce a discretized interacting-particle sampler. The leading interaction term is computed efficiently using vector-Jacobian products, while the curvature term can be estimated via finite differences or alternative trace/Laplacian estimators.
\end{itemize}

\begin{remark}
    We present our theoretical results under the Variance Exploding (VE) SDE parameterization, as it yields considerably simpler formulas. 
    However, the obtained results readily generalize to other parametrizations (i.e. general linear diffusion models) such as VP-SDEs.
\end{remark} 

\paragraph{Notation.} 
We use boldface symbols to denote batches of particles. For instance, $\bm{x}=(x^{(1)},\ldots,x^{(n)})\in(\R^d)^n$ represents a batch of states, and $\bm{X}_t=(X_t^{(1)},\ldots,X_t^{(n)})$ denotes the batch trajectory at time $t$. For any function $f:\R^d\to\R^k$, applying it to a batch implies element-wise application: $f(\bm{x}):= \big(f(x^{(1)}),\ldots,f(x^{(n)})\big)$. This convention naturally extends to linear operators, where $A\bm{x}:= (Ax^{(1)},\ldots,Ax^{(n)})$, and to gradients, where $\grad_{\bm{x}}F(\bm{x}) := \big(\grad_{x^{(1)}}F(\bm{x}),\ldots,\grad_{x^{(n)}}F(\bm{x})\big)$.
\section{Background}
\label{sec:background}

\subsection{Diffusion Models}
Let $p_0$ be the target data distribution on $\R^d$. We consider a forward process over the interval $[0,T]$ starting at $p_0$ which gradually noises samples towards pure noise:
\begin{equation}
    dX_t=\sqrt{2}\,dB_t,\quad X_0 \sim p_0,
    \label{eq:ve_kernel}
\end{equation}
where $B_t$ is standard Brownian motion. 
The marginal distribution at time $t$ can be represented as $X_t = X_0 + \sqrt{2t}\varepsilon$, where $\varepsilon \sim \mathcal N(0,\Id_d)$. We denote the density of $X_t$ as $p_t$.

Under mild conditions~\cite{ANDERSON1982313}, the reverse process $Y_{t} := X_{T-t}$ satisfies the following SDE:
\begin{equation}
    dY_t=2s_{T-t}(Y_t)\,dt+\sqrt{2}\,d\bar B_t,
    \qquad Y_0\sim p_T,
    \label{eq:ve_reverse}
\end{equation}
where $s_t=\nabla \log p_t$ is the score function and $\bar B_t$ is another standard Brownian motion.

By simulating the backward process~\eqref{eq:ve_reverse}, one can draw a sample $Y_T\sim p_0$ from the target distribution. In practice, the score $s_t$ is replaced by a neural network approximation, the initial distribution $p_T$ is approximated by $\mathcal{N}(0, 2T \cdot \Id_d)$, and the backward dynamics~\eqref{eq:ve_reverse} are discretized. We refer readers to~\citet{karras2022elucidatingdesignspacediffusionbased} for further details.

\begin{figure*}[!t]
    \centering
    \includegraphics[width=\textwidth]{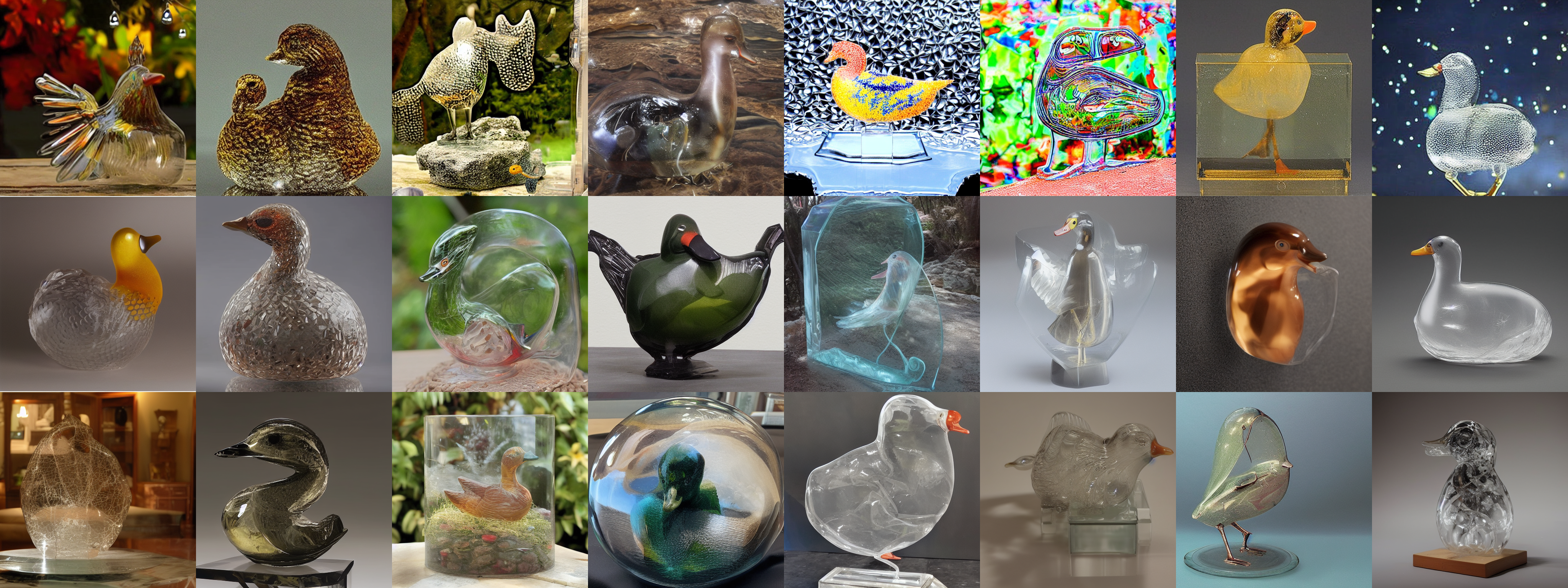}
    \caption{
    Qualitative comparison of diverse sampling for the prompt `\emph{A transparent sculpture of a duck made out of glass}' class. Rows correspond to \textbf{VT + Divergence} (ours, top), \textbf{VT} (ours, middle), and \textbf{CFG} (bottom).
    Our approach produces a more diverse set of  samples, covering a wider range of poses, backgrounds, and visual styles.}
    \label{fig:corgi_diverse_sampling}
\end{figure*}

\subsection{Terminal Reweighting and Doob Transforms}
\label{sec:doob_transform}
Suppose that a target distribution $q_0$ is obtained by tilting $p_0$ by the function $f$:
\begin{equation}
    q_0(x)\propto f(x)p_0(x),
    \qquad f(x)\ge 0.
\end{equation}
Let $q_t$ denote the marginal distribution at time $t$ obtained by running the same forward process~\eqref{eq:ve_kernel} starting from $q_0$. Then,
\begin{equation}
    \!\!\!\!q_t(x)\!\propto \!h_t(x)p_t(x)
    \text{ with }
    h_t(x)\!:=\!\mathbb E[f(X_0)\!\mid\! X_t=x].
    \label{eq:doob_scalar}
\end{equation}
Consequently, we can sample from $q_0$ by simulating the reverse process~\eqref{eq:ve_reverse} under the modified score $\nabla\log q_t$ where 
\begin{equation}
    \nabla\log q_t(x)=s_t(x)+\nabla\log h_t(x).
    \label{eq:score_doob_scalar}
\end{equation}
We refer to $h_t$ as the Doob $h$-transform for the tilt $f$.
We see that this provides the exact score correction to $s_t$ induced by the terminal weight $f$. We refer to \citet{denker2025deftefficientfinetuningdiffusion} for further details.

Crucially for our method, this formulation naturally extends to joint distributions over batches. Starting from the independent batch distribution $p_0^{\otimes n}$ and tilting by a non-negative function $f(\bm{x})$ yields the modified batch score:
\begin{equation}
    \nabla_{\bm{x}}\log q_t(\bm{x})
    =
    s_t(\bm{x})+
    \nabla_{\bm{x}}\log h_t(\bm{x}),
    \label{eq:batch_doob_score}
\end{equation}
where the batch Doob $h$-transform is defined as:
\begin{equation}
    h_t(\bm{x})
    :=
    \mathbb E[f(\bm{X}_0)\mid \bm{X}_t=\bm{x}].
    \label{eq:batch_doob_h}
\end{equation}
The remainder of this paper is dedicated to identifying this exact quantity for the setting of the variance-weighted distribution \eqref{eq:target_intro}.
\section{Main Result}
\label{sec:variance_target}
In this paper, we propose sampling from the following reweighting of the target distribution to encourage batch diversity: 
\begin{equation}
    \pi_0^A(\bm{x})
    \propto
    \Var_n^A(\bm{x})\prod_{i=1}^n p_0(x^{(i)}),
    \label{eq:pi_A}
\end{equation}
where we assume access to an approximation of the score $s_t$ for $p_0$, $A:\mathbb R^d\to\mathbb R^k$ is a linear operator projecting the images to the feature space in which we want to encourage diversity and $\Var_n^A(\bm{x})$ is the empirical variance of the features, namely 
\begin{equation}
    \Var_n^A(\bm{x})
    :=
    \frac1n\sum_{i=1}^n
    \norm{A(x^{(i)}-\bar x)}^2, \text{ where } \bar{x} = \frac{1}{n}\sum_{i=1}^n x^{(i)}.
\end{equation}

This is not a product distribution. It favors batches whose elements remain likely under the original model while being collectively spread out after applying $A$. Note that we enforce diversity directly at the data space $x_0$, compared to~\citet{corso2023particleguidancenoniiddiverse} which forces diversity on the noised latents $x_t$.

As discussed in Section~\ref{sec:doob_transform}, in order to sample from this distribution, we need to estimate the Doob $h$-transform associated with \eqref{eq:pi_A}
\begin{equation}
    h_t^A(\bm{x}_t)
    :=
    \mathbb E\left[\Var_n^A(\bm{X}_0)\mid \bm{X}_t=\bm{x}_t\right].
    \label{eq:h_A_def}
\end{equation}
Once $h_t^A$ is available, the exact target score is
\begin{equation}
    s_t^{\pi^A}(\bm{x}_t)
    =
    s_t(\bm{x}_t)+\nabla_{\bm{x}_t}\log h_t^A(\bm{x}_t).
    \label{eq:target_score_A}
\end{equation}
The target sampler is obtained from the usual reverse diffusion by replacing the original score with \eqref{eq:target_score_A}.

In the next section, we show how to compute the correction analytically. 
\subsection{Closed-Form Doob Correction}
\label{sec:closed_form}
We start by recalling Tweedie's identities, connecting the score and its derivatives with the moments of the posterior distribution:
\begin{equation}
    \mu_t(x):=\mathbb E[X_0\mid X_t=x]
    =x+2t\,s_t(x),
    \label{eq:ve_mu}
\end{equation}
and
\begin{equation}
    \Sigma_t(x):=\Cov(X_0\mid X_t=x)
    =2t\Id_d+4t^2\nabla s_t(x),
    \label{eq:ve_sigma}
\end{equation}
where $\nabla s_t$ denotes the Jacobian of the score function.
We provide their derivations in Appendix~\ref{app:tweedie}.

\begin{figure*}[!t]
    \centering
    \includegraphics[width=\textwidth]{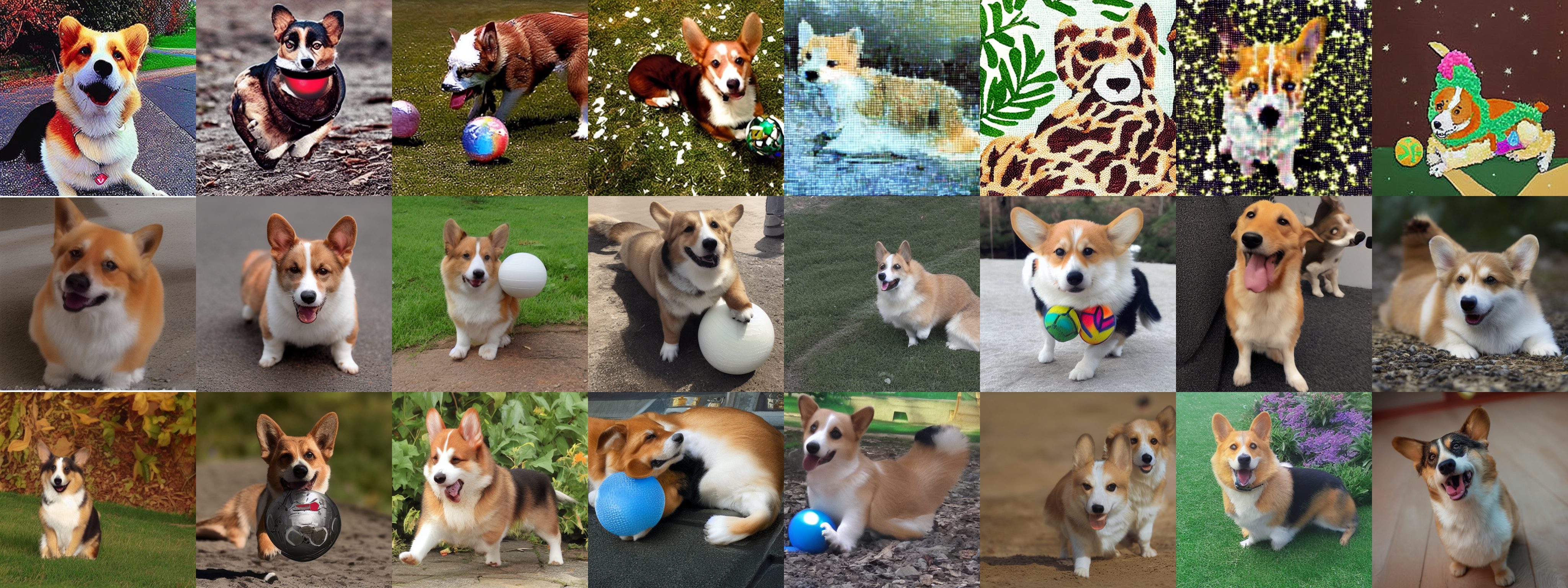}
    \caption{
    Qualitative comparison of diverse sampling for the prompt \emph{corgi with a ball}. Rows correspond to \textbf{VT + Divergence} (ours, top), \textbf{VT} (ours, middle), and \textbf{CFG} (bottom).
    Our approach produces a more diverse set of  samples, covering a wider range of poses, backgrounds, and visual styles.}
    \label{fig:duck_diverse_sampling}
\end{figure*}

Since $h_t^A$ is the expectation of the batch variance, by applying a bias-variance decomposition (see Appendix~\ref{app:bias_variance} for details), we can express $h_t^A$ in terms of the posterior mean $\mu_t$ and the posterior covariance $\Sigma_t$ in the following way
\£
\label{eq:h_A_general}
    &h_t^A(\bm{x}_t) 
    = \E\left[\Var_n^A(\bm{X}_0)\mid \bm{X}_t=\bm{x}_t\right]
    \\
    &\!\!\!\!=\!\Var_n\del{\E\left[A\bm{X}_0\!\mid\! \bm{X}_t=\bm{x}_t\right]} \!+\! \frac{n-1}{n^2}\!\sum_{i=1}^n\!\Var[AX_0|X_t\!=\!x^{(i)}_t] \nonumber
\£
This decomposition represents $h_t^A$ as a sum of two tractable terms.
The former rewards the repelling of the posterior means of the particles, while the latter rewards the particles for being in the region where the posterior distribution of features is diverse.

Substituting ${\Var[AX_0|X_t\!=\!x^{(i)}_t] =\Tr\left(\!A\Sigma_t(x_t^{(i)})A^\top \!\right)}$ and Tweedie's identities \eqref{eq:ve_mu} and \eqref{eq:ve_sigma} into~\eqref{eq:h_A_general} we rewrite $h_t^A$ in terms of the score function and its derivatives
\begin{align}
    \label{eq:h_A_ve}
    &h_t^A(\bm{x}_t) =
    \Var_n^A(\bm{x}_t+2t\,s_t(\bm{x}_t))\\
    &\quad+
    \frac{n-1}{n^2}
    \sum_{i=1}^n
    \left[
    2t\Tr(A^\top A)+4t^2\Tr\left(A\nabla s_t(x_t^{(i)})A^\top \right)
    \right]. \notag
\end{align}

Finally, the guidance term of interest in~\eqref{eq:target_score_A} has the form
\begin{equation}
    \nabla_{\bm{x}_t}\log h_t^A(\bm{x}_t) = \frac{\grad h_t^A(\bm{x}_t)}{h_t^A(\bm{x}_t)}
    =
    \frac{\big(g_t^{A,1}(\bm{x}_t),\ldots,g_t^{A,n}(\bm{x}_t)\big)}{h_t^A(\bm{x}_t)},
    \label{eq:grad_log_h_A}
\end{equation}
where 
\begin{align}
    g_t^{A,i}(\bm{x}_t)
    & =
    \frac{2}{n}
    J_{\mu,t}(x_t^{(i)})^\top
    A^\top A\left(\mu_t(x_t^{(i)})-\bar\mu_t(\bm{x}_t)\right)
    \notag\\
    &\quad+
    \frac{n-1}{n^2}4t^2
    \Div\left(A^\top A\nabla s_t(x_t^{(i)})\right)
    \label{eq:g_A}
\end{align}
and
\[
J_{\mu,t}(x):=I_d+2t\nabla s_t(x), \quad \bar{\mu}_t(\bm{x}_t) = \frac{1}{n}\sum_{i=1}^n \mu(x^{(i)}_t).
\]
We note that the divergence in~\eqref{eq:grad_log_h_A} is interpreted column-wise.

\begin{remark}[Projection specialization.]
When $P=A^\top A$ is the orthogonal projection onto a subspace
$U\subseteq \mathbb R^d$, let $a_1,\ldots,a_k$ denote the orthonormal
basis of $U$ from the rows of $A$. Then the divergence term in \eqref{eq:g_A} is the Laplacian of $s_t$ projected onto $U$, i.e.
\[
\Div\left(A^TA\nabla s_t(x)\right) = 
\sum_{\ell=1}^k \partial_{a_\ell}^2 s_t(x) =: \Delta_U s_t(x),
\]
where $\partial_{a_\ell}$ denotes the derivative in the
direction $a_\ell$.
For further details, see Appendix \ref{app:proj}.
\end{remark}
\subsection{Practical Implementation}
\label{sec:practical}

We sample from \eqref{eq:pi_A} by discretizing the reverse-time SDE and using the target score \eqref{eq:target_score_A}. At a reverse time corresponding to noise level $t$, each particle receives the usual score $s_t(x_t^{(i)})$ plus the Doob correction \eqref{eq:grad_log_h_A}. The leading term in $g_t^{A,i}$ can be computed efficiently with a vector-Jacobian product through the denoiser $\mu_t$. The divergence term requires computation of the second-order derivatives of $s_t$; in our implementation, we estimate it by finite differences. Other estimators, including Hutchinson-type stochastic trace methods or automatic-differentiation-based higher-order derivatives, can be substituted.
\begin{algorithm}[tb]
\caption{Variance-weighted diffusion sampling}
\label{alg:exact_diverse_diffusion}
\begin{algorithmic}[1]
\STATE \textbf{Input:} score model $s_t$, linear map $A$, batch size $n$, time grid $T=t_K>\cdots>t_0=0$, scaling term $\lambda>0$ 
\STATE Initialize $\bm{x}_{t_K}=(x_{t_K}^{(1)},\ldots,x_{t_K}^{(n)})$ by sampling independently from $\cN\del{0, 2T\Id_d}$
\FOR{$k=K,K-1,\ldots,1$}
    \STATE Set $t=t_k$ and compute $\mu_t(x_{t}^{(i)})=x_t^{(i)}+2t s_t(x_t^{(i)})$ for all particles
    \STATE Compute $\bar\mu_t=n^{-1}\sum_{i=1}^n\mu_t(x_t^{(i)})$
    \STATE Compute~\eqref{eq:h_A_ve} to get $h_t^A(\bm{x}_t)$ using VJP  \FOR{$i=1,\ldots,n$}
        \STATE Set $v_i=A^\top A(\mu_t(x_t^{(i)})-\bar\mu_t)$
        \STATE Compute $r_i=J_{\mu,t}(x_t^{(i)})^\top v_i$ using VJP
        \STATE Estimate $c_i\approx \Div(A^\top A\nabla s_t(x_t^{(i)}))$ using finite differences
        \STATE Set $g_i=\frac{2}{n}r_i+\frac{n-1}{n^2}4t^2 c_i$
        \STATE Set $\widehat s_i=s_t(x_t^{(i)})+\lambda g_i/h_t^A(\bm{x}_t)$
    \ENDFOR
    \STATE Take one reverse SDE or ODE discretization step using scores $(\widehat s_1,\ldots,\widehat s_n)$
\ENDFOR
\STATE \textbf{Return:} $\bm{x}_{t_0}$
\end{algorithmic}
\end{algorithm}

\section{Experiments}
\label{sec:experiments}

We evaluate our approach for image generation. 
In particular, we take Stable Diffusion v1.5 at 512x512 \citep{Rombach_2022_CVPR} and apply our variance-tilted Doob $h$-transform with the score and posterior mean estimator obtained from classifier-free guidance (CFG) \citep{ho2022classifier} for a variety of different prompts.
We test (i) the variance-tilted Doob $h$-transform which includes the estimated divergence term and (ii) without the divergence term (i.e. we drop the $\Tr(A^\top A \nabla s_t)$ term from $h_t^A$) against the standard CFG baseline where the initial noise for each approach is set to be the same.
We also set $A=\Id$.
For further details, see Appendix \ref{app:run_512_cfg}.

We present qualitative results in Figures \ref{fig:corgi_diverse_sampling} and \ref{fig:duck_diverse_sampling}. Specifically, we compare the standard \textbf{CFG} baseline against our two proposed configurations: Variance-Tilted (\textbf{VT}) without the divergence term, and \textbf{VT + Divergence}, which includes the estimated divergence term. These initial findings are promising and suggest that our variance--weighted distribution achieves meaningful diversity. Rather than simply producing minor variations of high--probability modes, our approach successfully yields a more diverse set of samples that cover a significantly wider range of poses, backgrounds, and visual styles.

\section{Discussion and Conclusion}
\label{sec:discussion_conclusion}

We proposed a target-first approach to diverse diffusion sampling. By reweighting independent samples according to empirical variance after a linear feature map, we obtain an explicit batch distribution and derive its exact Doob correction under Gaussian noising. The resulting sampler has a clear decomposition into denoised-mean repulsion and a curvature correction. This provides a probabilistic alternative to heuristic particle repulsion, while retaining practical implementation paths based on vector-Jacobian products and numerical estimators for the curvature term.

\paragraph{Limitations.} 

While our method qualitatively produces more diverse outputs, it can negatively affect image quality somewhat. Future work could explore how to anneal the Doob correction term to mitigate these effects, as well as quantitatively measuring the performance of our method. Additionally, when including the full divergence term, our method is more expensive than standard sampling from diffusion models due to the cost of estimating the divergence.
Future work could explore approaches for reducing this cost while retaining diverse sampling.
\section*{Acknowledgments}
IA is supported by the Engineering and Physical Sciences Research Council [grant number EP/T517811/1]. 
LZ and KA are supported by the EPSRC CDT in Modern Statistics and Statistical Machine Learning (EP/S023151/1).

\section*{Impact Statement}

This paper presents work whose goal is to advance the field of Machine Learning. There are many potential societal consequences of our work, none of which we feel must be specifically highlighted here.

\bibliographystyle{apalike}
\bibliography{example_paper}

\clearpage
\onecolumn
\appendix

\section{Doob $h$-Transform for Terminal Reweighting}
\label{app:doob_transform}

We first record the identity underlying the construction. Let $P$ denote the law of the forward process initialized from $p_0$, and let $Q$ be the path measure obtained by reweighting the initial point by a nonnegative function $f$:
\begin{equation}
    \frac{dQ}{dP}(X_{0:T})
    =
    \frac{f(X_0)}{\mathbb E_P[f(X_0)]}.
\end{equation}
Let $p_t$ and $q_t$ denote the corresponding time-$t$ marginals. For any test function $\varphi$,
\begin{align}
    \mathbb E_Q[\varphi(X_t)]
    & =
    \frac{\mathbb E_P[\varphi(X_t)f(X_0)]}{\mathbb E_P[f(X_0)]}
    \notag\\
    & =
    \frac{\mathbb E_P[\varphi(X_t)\mathbb E_P[f(X_0)\mid X_t]]}{\mathbb E_P[f(X_0)]}.
\end{align}
Hence
\begin{equation}
    q_t(x)
    =
    \frac{h_t(x)p_t(x)}{\mathbb E_P[f(X_0)]},
    \qquad
    h_t(x):=\mathbb E_P[f(X_0)\mid X_t=x].
\end{equation}
Assuming differentiability and positivity of $p_t$ and $h_t$, this gives
\begin{equation}
    \nabla\log q_t(x)
    =
    \nabla\log p_t(x)+\nabla\log h_t(x).
\end{equation}
The same argument applies to the batch process $\bm X_t=(X_t^{(1)},\ldots,X_t^{(n)})$ initialized from $p_0^{\otimes n}$ with $f$ replaced by any nonnegative batch functional $f(\bm x)$.

\section{Conditional Moments for Linear-Gaussian Corruption}
\label{app:tweedie}

Consider a linear-Gaussian noising model
\begin{equation}
    X_t=\alpha_t X_0+\sigma_t\varepsilon,
    \qquad
    \varepsilon\sim\mathcal N(0,I_d),
    \qquad
    \alpha_t>0,
    \quad
    \sigma_t>0.
    \label{eq:app_linear_gaussian}
\end{equation}
Let $p_t$ be the density of $X_t$ and $s_t(x)=\nabla_x\log p_t(x)$. The conditional density of $X_t$ given $X_0=x_0$ is Gaussian with covariance $\sigma_t^2I_d$. Differentiating the marginal density gives
\begin{equation}
    \nabla_x p_t(x)
    =
    \int -\frac{x-\alpha_t x_0}{\sigma_t^2}
    p_{t|0}(x\mid x_0)p_0(x_0)\,dx_0.
\end{equation}
Dividing by $p_t(x)$ yields
\begin{equation}
    s_t(x)
    =
    -\frac{1}{\sigma_t^2}
    \left(x-\alpha_t\mathbb E[X_0\mid X_t=x]\right).
\end{equation}
Therefore the posterior mean is
\begin{equation}
    \mu_t(x):=\mathbb E[X_0\mid X_t=x]
    =
    \frac{1}{\alpha_t}\left(x+\sigma_t^2s_t(x)\right).
    \label{eq:app_mu_general}
\end{equation}
Differentiating \eqref{eq:app_mu_general} gives
\begin{equation}
    J_{\mu,t}(x):=\nabla_x\mu_t(x)
    =
    \frac{1}{\alpha_t}\left(\Id_d+\sigma_t^2\nabla s_t(x)\right).
    \label{eq:app_jmu_general}
\end{equation}
The posterior covariance follows from the second-derivative identity
\begin{equation}\label{eq:grad-mu}
    \nabla_x\mu_t(x)
    =
    \frac{\alpha_t}{\sigma_t^2}\Cov(X_0\mid X_t=x),
\end{equation}
which is obtained by differentiating the posterior expectation under the Gaussian likelihood. Combining this identity with \eqref{eq:app_jmu_general} gives
\begin{equation}
    \Sigma_t(x):=\Cov(X_0\mid X_t=x)
    =
    \frac{\sigma_t^2}{\alpha_t^2}I_d
    +
    \frac{\sigma_t^4}{\alpha_t^2}\nabla s_t(x).
    \label{eq:app_sigma_general}
\end{equation}

\section{Bias-Variance Decomposition for the Batch Variance}
\label{app:bias_variance}

We now justify the decomposition used to compute the Doob term. Condition on $\bm X_t=\bm x_t$. Define
\begin{equation}
    Z_i:=AX_0^{(i)},
    \qquad
    \bar Z:=\frac1n\sum_{j=1}^n Z_j,
\end{equation}
and
\begin{equation}
    \mu_i^A
    :=
    \mathbb E[Z_i\mid \bm X_t=\bm x_t]
    =
    A\mu_t(x_t^{(i)}),
    \qquad
    \Sigma_i^A
    :=
    \Cov(Z_i\mid \bm X_t=\bm x_t)
    =
    A\Sigma_t(x_t^{(i)})A^\top.
\end{equation}
Since the forward noising is independent across particles, $Z_1,\ldots,Z_n$ are conditionally independent given $\bm X_t=\bm x_t$. The empirical variance can be written as
\begin{equation}
    \frac1n\sum_{i=1}^n\norm{Z_i-\bar Z}^2
    =
    \frac1n\sum_{i=1}^n\norm{Z_i}^2
    -
    \norm{\bar Z}^2.
    \label{eq:app_emp_var_identity}
\end{equation}
For the first term,
\begin{equation}
    \mathbb E\left[\norm{Z_i}^2\mid \bm X_t=\bm x_t\right]
    =
    \norm{\mu_i^A}^2+
    \Tr(\Sigma_i^A).
    \label{eq:app_first_moment_z}
\end{equation}
For the second term, conditional independence gives
\begin{equation}
    \mathbb E\left[\norm{\bar Z}^2\mid \bm X_t=\bm x_t\right]
    =
    \norm{\bar\mu^A}^2
    +
    \frac{1}{n^2}\sum_{i=1}^n\Tr(\Sigma_i^A),
    \label{eq:app_bar_z_second}
\end{equation}
where
\begin{equation}
    \bar\mu^A:=\frac1n\sum_{i=1}^n\mu_i^A.
\end{equation}
Substituting \eqref{eq:app_first_moment_z} and \eqref{eq:app_bar_z_second} into \eqref{eq:app_emp_var_identity}, we obtain
\begin{align}
    \mathbb E\left[
    \frac1n\sum_{i=1}^n\norm{Z_i-\bar Z}^2
    \middle|\bm X_t=\bm x_t
    \right]
    & =
    \frac1n\sum_{i=1}^n\norm{\mu_i^A}^2
    -
    \norm{\bar\mu^A}^2
    \notag\\
    &\quad+
    \left(\frac1n-\frac1{n^2}\right)
    \sum_{i=1}^n\Tr(\Sigma_i^A).
\end{align}
The first line is the empirical variance of the conditional means,
\begin{equation}
    \frac1n\sum_{i=1}^n\norm{\mu_i^A}^2
    -
    \norm{\bar\mu^A}^2
    =
    \frac1n\sum_{i=1}^n\norm{\mu_i^A-\bar\mu^A}^2.
\end{equation}
Moreover,
\begin{equation}
    \frac1n-\frac1{n^2}
    =
    \frac{n-1}{n^2}.
\end{equation}
Therefore
\begin{equation}
    \mathbb E\left[\Var_n^A(\bm X_0)\mid \bm X_t=\bm x_t\right]
    =
    \Var_n^A(\mu_t(\bm x_t))
    +
    \frac{n-1}{n^2}
    \sum_{i=1}^n
    \Tr\left(A\Sigma_t(x_t^{(i)})A^\top\right).
    \label{eq:app_bias_variance_final}
\end{equation}
Using cyclicity of the trace and $B:=A^\top A$, this is equivalently
\begin{equation}
    \mathbb E\left[\Var_n^A(\bm X_0)\mid \bm X_t=\bm x_t\right]
    =
    \Var_n^A(\mu_t(\bm x_t))
    +
    \frac{n-1}{n^2}
    \sum_{i=1}^n
    \Tr\left(B\Sigma_t(x_t^{(i)})\right).
    \label{eq:app_h_general_proved}
\end{equation}

\section{Gradient of the Doob $h$-Transform}
\label{app:gradient_doob}

We compute the gradient of $h_t^A$ for a generic linear diffusion $X_t=\alpha_t X_0 + \sigma_t\epsilon$.

Let $B=A^\top A$, and write the Doob $h$-transform as
\begin{equation}
    h_t^A(\bm x_t)
    =
    \Var_n^A(\mu_t(\bm x_t))
    +
    \frac{n-1}{n^2}\sum_{i=1}^n
    \Tr\left(B\Sigma_t(x_t^{(i)})\right).
\end{equation}
To compute the gradient of the first term, we first define
\begin{equation}
    \bar\mu_t(\bm x_t):=\frac1n\sum_{j=1}^n\mu_t(x_t^{(j)}).
\end{equation}
We now compute the gradient by
\begin{align}
    \nabla_{x_t^{(i)}}\Var_n^A(\mu_t(\bm x_t)) 
    &= \nabla_{x_t^{(i)}} \frac{1}{n}\sum_{j=1}^n \norm{A\mu_t(x_t^{(j)}) - A\bar\mu_t(\bm x_t) }^2 \\
    &= \nabla_{x_t^{(i)}} \frac{1}{n} \sum_{j=1}^n \norm{A\mu_t(x_t^{(j)})}^2 - \nabla_{x_t^{(i)}} \norm{A\bar\mu_t(\bm x_t)}^2 \\
    &= \frac{2}{n} \left( AJ_{\mu, t}(x_t^{(i)}) \right)^\top A\mu_t(x_t^{(i)}) - 2\left( A J_{\bar\mu, t}(x_t^{(i)})\right)^\top A\bar\mu_t(x_t).
\end{align}
where $J_{\mu, t}, J_{\bar\mu, t}$ denote the Jacobian of $\mu_t$ and $\bar\mu_t$ with respect to $x_t^{(i)}$ respectively.
From~\eqref{eq:grad-mu}, we have
\begin{align}
    J_{\mu,t}(x_t^{(i)})
    =
    \frac{1}{\alpha_t}\left(\Id_d+\sigma_t^2\nabla s_t(x_t^{(i)})\right),
\end{align}
and by the definition of $\bar\mu_t$, we have
\begin{align}
    J_{\bar\mu, t}(x_t^{(i)}) = \frac{1}{n} J_{\mu, t}(x_t^{(i)}).
\end{align}
Putting this together, we obtain for the gradient of the first term:
\begin{align}
     \nabla_{x_t^{(i)}}\Var_n^A(\mu_t(\bm x_t)) 
     &= \frac{2}{n} J_{\mu, t}(x_t^{(i)})^\top B \left( \mu_t(x_t^{(i)}) - \bar\mu_t(x_t^{(i)}) \right).
\end{align}

Regarding the second term, we have
\begin{equation}
    \nabla_{x_t^{(i)}}
    \sum_{j=1}^n\Tr\left(B\Sigma_t(x_t^{(j)})\right)
    =
    \nabla_{x_t^{(i)}}\Tr\left(B\Sigma_t(x_t^{(i)})\right).
    \label{eq:app_grad_second_term_general}
\end{equation}
Using \eqref{eq:app_sigma_general}, we have
\begin{equation}
    \Tr(B\Sigma_t(x_t^{(i)}))
    =
    \frac{\sigma_t^2}{\alpha_t^2}\Tr(B)
    +
    \frac{\sigma_t^4}{\alpha_t^2}\Tr(B\nabla s_t(x_t^{(i)})).
\end{equation}
The first term is independent of $x_t^{(i)}$, so
\begin{equation}
    \nabla_{x_t^{(i)}}\Tr(B\Sigma_t(x_t^{(i)}))
    =
    \frac{\sigma_t^4}{\alpha_t^2}
    \nabla_{x_t^{(i)}}\Tr(B\nabla s_t(x_t^{(i)})).
    \label{eq:app_grad_trace_general}
\end{equation}
Moreover, as the above expression is a linear combination of the third derivatives of $\log p_t$ and they commute, we have
\begin{align}
    \nabla_{x_t^{(i)}}\Tr(B\Sigma_t(x_t^{(i)}))
    = \frac{\sigma_t^4}{\alpha_t^2}
    \Div(B\nabla s_t(x_t^{(i)})),
\end{align}
where the divergence is applied column-wise.
To see this, we look at the $k$-th component of $\nabla_x \Tr(B\nabla s_t(x))$:
\begin{align}
    \partial_{x_k} \Tr(B\nabla s_t(x))
    &= \partial_{x_k} \sum_{j} (B\nabla s_t(x))_{jj} \\
    &= \partial_{x_k} \sum_{j, l} B_{jl} [\nabla s_t(x)]_{lj} \\
    &=  \sum_{j, l} B_{jl} \partial_{x_k}\partial_{x_l}\partial_{x_j} \log p_t(x),
\end{align}
and the $k$-th component of $\Div(B\nabla s_t(x))$:
\begin{align}
    \Div(B\nabla s_t(x))_k 
    &= \sum_{l} \partial_{x_l} (B\nabla s_t(x))_{lk} \\
    &= \sum_{l} \partial_{x_l} \sum_{j} B_{lj} [\nabla s_t(x)]_{jk} \\
    &= \sum_{l, j}  B_{lj} \partial_{x_l}\partial_{x_j}\partial_{x_k} \log p_t(x).
\end{align}
As the derivatives of $\log p_t$ commute, we can conclude.

\section{Projection Specialization}\label{app:proj}

We consider the $i$-th component of the LHS of the identity:
\begin{align}
    \Div(A^\top A\nabla s_t(x))_i 
    &= \sum_{\ell} \partial_{x_\ell} (A^\top A\nabla s_t(x))_{\ell i} \\
    &= \sum_\ell \partial_{x_\ell} \sum_{j} (A^\top A)_{\ell j}[\nabla s_t(x)]_{ji} \\
    &= \sum_{\ell, j} (A^\top A)_{\ell j} \partial_{x_\ell} \partial_{x_i} (s_{t})_j(x) \\
    &= \sum_{\ell, j} (A^\top A)_{\ell j} \partial_{x_\ell} \partial_{x_j} (s_{t})_i(x) \\
\end{align}
where we recall the divergence is considered column-wise and the final equality holds as the derivatives of $\log p_t$ commute.
Furthermore, the $i$-th component of the RHS of the identity satifies:
\begin{align}
    \sum_{\ell} \partial^2_{a_\ell} (s_t)_i(x) 
    &= \sum_\ell \partial_{a_\ell} \sum_{j} (a_\ell)_j \partial_{x_j} (s_t)_i(x) \\
    &= \sum_{\ell, j} (a_\ell)_j \partial_{a_\ell} [\partial_{x_j} (s_t)_i(x)] \\
    &= \sum_{\ell, j} (a_\ell)_j \sum_{k} (a_\ell)_k \partial_{x_k} \partial_{x_j} (s_t)_i(x) \\
    &= \sum_{j, k}  \left[\sum_{\ell} (a_\ell)_j (a_\ell)_k \right]  \partial_{x_k} \partial_{x_j} (s_t)_i(x) \\
    &= \sum_{j, k}  (A^\top A)_{jk}  \partial_{x_k} \partial_{x_j} (s_t)_i(x),
\end{align}
where the last equality holds as $a_\ell$ is the $\ell$-th row of $A$.
Finally, we see that the two sides are equal from the fact $A^\top A$ is symmetric and the derivatives of $\log p_t$ commute.

\section{Coordinate Masks}
\label{app:coordinate_masks}

When $A$ is a coordinate mask, $B=A^\top A$ is diagonal. Writing $B=\diag(b)$,
\begin{equation}
    \Tr(B\nabla s_t(x))
    =
    \sum_{a=1}^d b_a\partial_{x_a}s_{t,a}(x)
    =
    \sum_{a=1}^d b_a\partial_{x_a}^2\log p_t(x).
\end{equation}
Thus the trace term is the masked scalar Laplacian of $\log p_t$. Its gradient is the corresponding coordinatewise vector Laplacian of the score,
\begin{equation}
    \nabla_x\Tr(B\nabla s_t(x))
    =
    \sum_{a=1}^d b_a\partial_{x_a}^2s_t(x).
\end{equation}
For binary masks, masking by $A$ and masking by $B=A^\top A$ coincide. For soft masks, one should specify whether the intended weights are those in $A$ or those in $A^\top A$.

\section{Further Details on Experiments}
\label{app:run_512_cfg}

This section describes our empirical implementation.

\paragraph{Base CFG score.}

We use the classifier-free guidance score term $s_{\mathrm{cfg}}$ defined by 
\[
s_{\mathrm{cfg}}(x_t,t)
=
s_{\mathrm{cond}}(x_t,t)
+w\big(s_{\mathrm{cond}}(x_t,t)-s_{\mathrm{uncond}}(x_t,t)\big),
\]
where $w$ is some scaling factor and $s_{\mathrm{cond}}$ denotes the score conditioned on some prompt and $s_{\mathrm{uncond}}$ denotes the unconditional score.
For all of our experiments, we set this to 1.

\paragraph{Doob $h$-transform.}
We compute the posterior mean considered under the CFG score $s_{\mathrm{cfg}}$ where we use Tweedie's formula.
Similar to CFG, we also scale the Doob correction $\nabla \log h_t^A$ with a controllable parameter $\lambda$; for all of our experiments we set this to 0.5.
For the first term in $g_t^{A, i}$ involving $J_{\mu, t}^\top$, we compute this with a vector-Jacobian product.
We include the second term $\frac{\sigma_t^4}{\alpha_t^2}\Div(B\nabla s_t(x_t^{(i)}))$ if we are including the divergence term in the correction. 
Additionally, even if the divergence term is included, we only include it if $\sigma_t^4/\alpha_t^2 \le c_\text{cutoff}$ where we set $c_\text{cutoff}$ to be 16 in all of our experiments to prevent instabilities.
Furthermore, we evaluate $\frac{\sigma_t^4}{\alpha_t^2}\Div(B\nabla s_t(x_t^{(i)}))$ with a Hutchinson estimator with $L=8$ orthogonalized Gaussian
probes.  Concretely, the implementation draws a Gaussian matrix in the
flattened batch-feature space, applies a QR factorization, rescales the
orthonormal columns by the square root of the feature dimension, reshapes each
probe to the batch feature shape, and lifts it back to latent space through
$A^\top$.  Thus each latent probe $u_\ell$ satisfies
$\mathbb{E}[u_\ell u_\ell^\top]=B$, while different probes are orthogonalized
within the same estimator call.
Additionally, we estimate the divergence term in $g_t^{A, i}$ with finite differences.

\section{Additional Results}
\label{app:additional_results}

\begin{figure*}[!h]
    \centering
    \includegraphics[width=\textwidth]{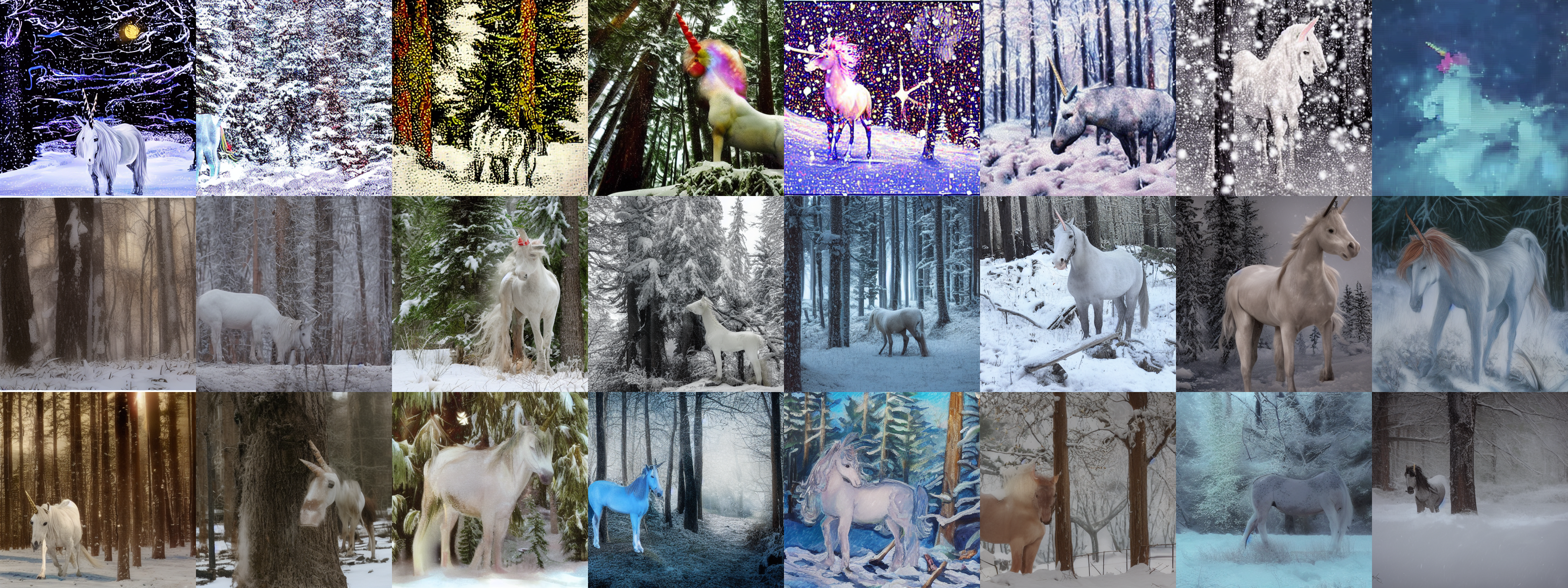}
    \caption{
    Qualitative comparison of diverse sampling for the prompt `\emph{A unicorn in a snowy forest}'. Rows correspond to \textbf{VT + Divergence} (ours, top), \textbf{VT} (ours, middle), and \textbf{CFG} (bottom).}
    \label{fig:unicorn_diverse_sampling}
\end{figure*}

\begin{figure*}[!h]
    \centering
    \includegraphics[width=\textwidth]{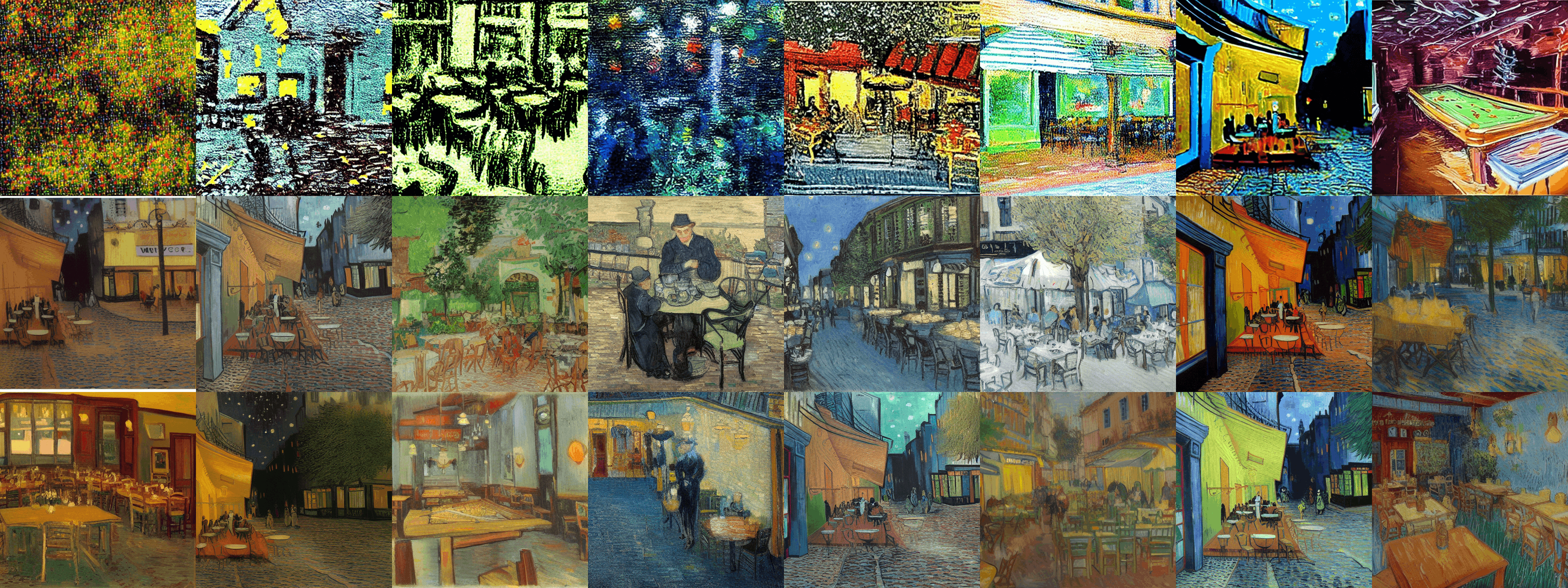}
    \caption{
    Qualitative comparison of diverse sampling for the prompt `\emph{Van Gogh Cafe Terasse}'. Rows correspond to \textbf{VT + Divergence} (ours, top), \textbf{VT} (ours, middle), and \textbf{CFG} (bottom).}
    \label{fig:gogh_diverse_sampling}
\end{figure*}

\begin{figure*}[!h]
    \centering
    \includegraphics[width=\textwidth]{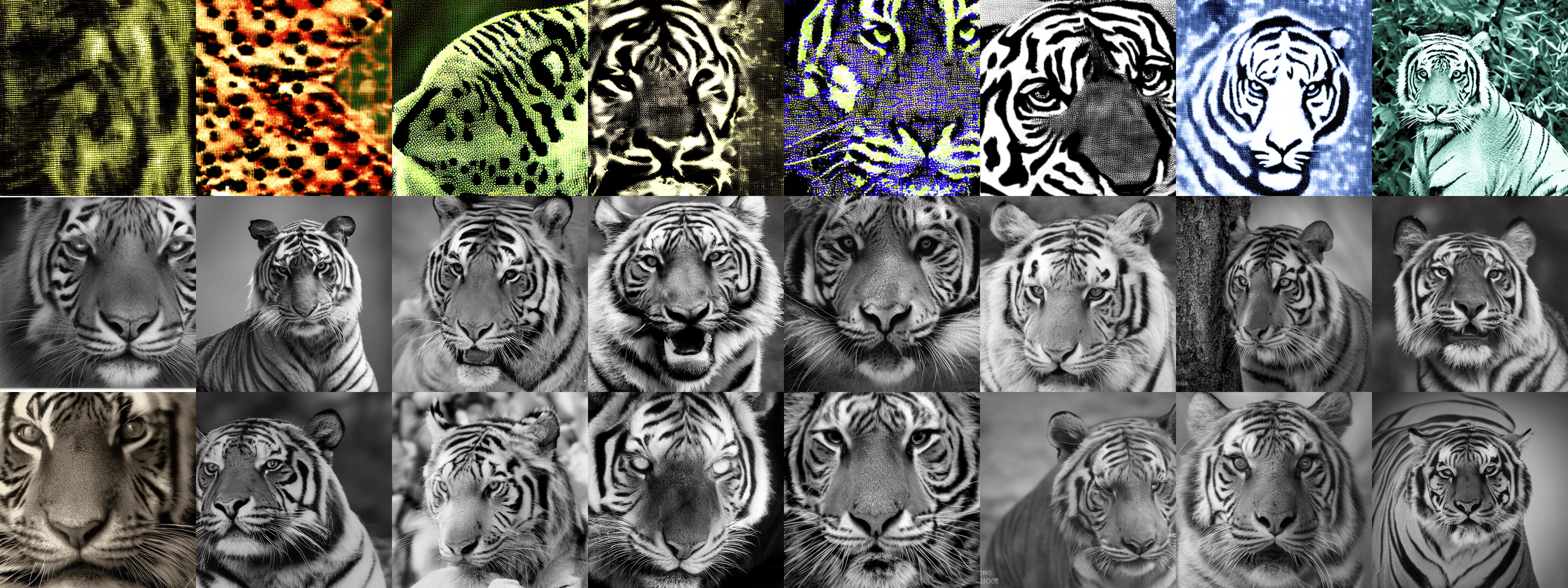}
    \caption{
    Qualitative comparison of diverse sampling for the prompt `\emph{Portrait of tiger in black and white by Lukas Holas}'. Rows correspond to \textbf{VT + Divergence} (ours, top), \textbf{VT} (ours, middle), and \textbf{CFG} (bottom).}
    \label{fig:tiger_diverse_sampling}
\end{figure*}

We present qualitative results for additional prompts in Figures \ref{fig:unicorn_diverse_sampling}-\ref{fig:tiger_diverse_sampling}.


\end{document}